\documentclass[
twocolumn,
]{ceurart}

\usepackage{listings}
\begin{document}

%%
%% Rights management information.
%% CC-BY is default license.
\copyrightyear{2022}
\copyrightclause{Copyright for this paper by its authors.
  Use permitted under Creative Commons License Attribution 4.0
  International (CC BY 4.0).}

%%
%% This command is for the conference information
\conference{Proceedings of the CIKM 2022 Workshops, October 17 - 22, 2022}
%CIKM'22: Workshop on Advances in Interpretable Machine Learning and Artificial Intelligence Workshop, 2022, Atlanta, Georgia, USA}

%%
%% The "title" command
\title{Interpretability in Activation Space Analysis of Transformers: A Focused Survey}

%\tnotemark[1]
%\tnotetext[1]{You can use this document as the template for preparing your
%  publication. We recommend using the latest version of the ceurart style.}

%%
%% The "author" command and its associated commands are used to define
%% the authors and their affiliations.
\author[1]{Soniya Vijayakumar}[%
%orcid=0000-0002-0877-7063,
email=soniya.vijayakumar@dfki.de,
%url=https://vksoniya.github.io/,
]
\cormark[1]
%\fnmark[1]
\address[1]{German Research Center for Artificial Intelligence (DFKI), Saarland Informatics Campus,
Saarland, Germany}
%\address[2]{Universität des Saarlandes,  Saarland Informatics Campus, Saarland, Germany}

%\author[3]{Ilaria Tiddi}[%
%orcid=0000-0001-7116-9338,
%email=i.tiddi@vu.nl,
%url=https://kmitd.github.io/ilaria/,
%]
%\fnmark[1]
%\address[3]{Vrije Universiteit Amsterdam, De Boelelaan 1105, 1081 HV Amsterdam, The Netherlands}

%\author[4]{Manfred Jeusfeld}[%
%orcid=0000-0002-9421-8566,
%email=Manfred.Jeusfeld@acm.org,
%url=http://conceptbase.sourceforge.net/mjf/,
%]
%\fnmark[1]
%\address[4]{University of Skövde, Högskolevägen 1, 541 28 Skövde, Sweden}

%% Footnotes
\cortext[1]{Corresponding author.}
%\fntext[1]{These authors contributed equally.}

%%
%% The abstract is a short summary of the work to be presented in the
%% article.
\begin{abstract}
   The field of natural language processing has reached breakthroughs with the advent of transformers. They have remained state-of-the-art since then, and there also has been much research in analyzing, interpreting, and evaluating the attention layers and the underlying embedding space. In addition to the self-attention layers, the feed-forward layers in the transformer are a prominent architectural component. From extensive research, we observe that its role is under-explored. We focus on the latent space, known as the \textit{Activation Space}, that consists of the neuron activations from these feed-forward layers. In this survey paper, we review interpretability methods that examine the learnings that occurred in this activation space. Since there exists only limited research in this direction, we conduct a detailed examination of each work and point out potential future directions of research. We hope our work provides a step towards strengthening activation space analysis.
\end{abstract}

%%
%% Keywords. The author(s) should pick words that accurately describe
%% the work being presented. Separate the keywords with commas.
\begin{keywords}
  explainability \sep interpretability \sep machine learning \sep
  activation space analysis \sep linguistic information \sep transformers \sep feed-forward layers
\end{keywords}

%%
%% This command processes the author and affiliation and title
%% information and builds the first part of the formatted document.
\maketitle

\section{Introduction}

Through thick and thin, there is evidence that transformers have established itself as the state-of-the-art in various Natural Language Processing (NLP) tasks since their conception and realization in 2017. BERT, the most well-known transformer language model \cite{DBLP:journals/corr/VaswaniSPUJGKP17}, consists of two major architectural components: self-attention layers and feed-forward layers. Much work has been done in analyzing the functions of self-attention layers \cite{Voitaetal, DBLP:journals/corr/abs-1906-04341, vig-belinkov-2019-analyzing}. In our survey, we focus on interpretability of the feed-forward layers. Each layer in the encoder and decoder contains a fully connected position-wise feed-forward network. The feed-forward network contains two linear transformations with a rectified linear activation function. Even though existing works highlight the importance of such feed-forward layers in transformers \cite{press-etal-2020-improving, https://doi.org/10.48550/arxiv.2105.05920, DBLP:journals/corr/abs-2007-06257}, still, to date, the role of feed-forward layers remains under-explored \cite{DBLP:journals/corr/abs-2104-08696}. Our review focuses on the research that uses interpretability methods to understand the learnings in these feed-forward layers. We define the latent space, that comprises of the activations extracted from these layers, as the \textit{Activation Space}. Many methods already exist for aggregating these representations including the default Huggingface\footnote{\url{https://huggingface.co/}} pipeline used in the original BERT paper \cite{DBLP:journals/corr/abs-1810-04805}. 

Several methods for explaining and interpreting deep neural networks have been devised and we observe that much of the focus is in the domain of image processing \cite{DBLP:journals/corr/abs-2006-11371}.
%\marginpar{SO: reference for a review article or such?}. 
A challenge that exists is the gap between the low-level features that the neural networks compute and the high-level concepts that are human-understandable. Furthermore, we observe that there have been relatively fewer research methods applied in understanding the internal learnings of networks in comparison to analyzing the functions of self-attention layers.

%\marginpar{SO: as compared to?} 
The core focus of our review is directed towards those methods that unfold the learnings in the internal representations of the neural network, i.e, we look at those methods that answer the question: ``What does the model learn?'' We further refine our focus on understanding specifically the feed-forward layers in transformer models. The motivation for this study is two-fold:
\begin{itemize}
    \item The inputs undergo a non-linear transformation when passing through the activation functions in the feed-forward layers of deep neural networks \cite{DBLP:journals/corr/abs-2101-04547}. %\marginpar{SO: Why is this interesting?}
    \item The parameters in the position-wise feed-forward layers of the transformer account for two-thirds of the total model's parameters ($8d^2$ per layer, \textit{d} is the model’s hidden dimension). This also implies that there is a considerable amount of computational budget involved in training these parameters to achieve the state-of-the-art performance they deliver today \cite{DBLP:journals/corr/abs-2012-14913}.
\end{itemize}

From recent research, the methods that focus on understanding the feed-forward layers show substantial evidence that the feed-forward layer activation space embeds useful information (see Section \ref{sec:aaa}). We find that the learnings in the feed-forward layer %even though it constitutes two-thirds of the transformer parameters, 
remain under-explored. %In our view, interpreting these learnings could help in parameter or architectural optimization. 
With our methodological survey, our objective is to understand the internal mechanisms of transformers by exploring the activation space of the feed-forward network. Further, we consider this paper as a focused starting point for facilitating future research in activation space analysis. Finally, we also conduct a comparative study of these methods, their evaluation techniques and report our observations, understandings, and potential future directions (see Section \ref{sec:ins}). Table \ref{tab:commands} summarizes the methods and its attributes that we have explored.

\begin{table*}[t]
  \caption{Major attributes of the methods explored in the activation space analysis methods}
    
  \label{tab:commands}
  %\begin{tabular}{p{4.5cm}p{4cm}p{3.5cm}p{3.5cm}}
  \begin{tabular}{p{2cm}p{2.6cm}p{2.6cm}p{2.7cm}p{2.6cm}}
  %\begin{tabular}{llll}
    
   \toprule
    \textbf{Method} 
    %& \textbf{Category} 
    & \textbf{Properties} 
    %&  \textbf{Generic Properties} 
    & \textbf{NLP Tasks} & \textbf{Quantitative Evaluation} & \textbf{Qualitative Evaluation}\\
    \midrule
     
     \textbf{Linguistic Phenomena}   \cite{dalvi2019neurox, durrani-etal-2021-transfer, DBLP:journals/corr/abs-2010-02695, alammar-2021-ecco} & 
     %Local (Individual Neurons), Post-hoc (Auxiliary Classifier) & 
     Word Morphology, Lexical Semantics, Sentence Length, Parts-of-Speech & 
     %Sentence Length & 
     Parts-of-Speech, Semantic and Syntax Tagging and Prediction, Syntactic Chunking & 
     Sensitivity, Prediction Accuracy, Selectivity Score
     & Human-expert visual inspection of selected neurons\\

    \textbf{Neural Memory Cells} \cite{DBLP:journals/corr/abs-2012-14913, DBLP:journals/corr/abs-2104-08696} & 
    %Global using Local (Individual Neurons) &
    %-- & 
    Vocabulary Distribution, Human-Interpretable Patterns, Factual Knowledge & 
    Next Sequence Prediction, Fill-in-the-blank Cloze Task &
    Agreement Rate, Prediction Probability, Attribution Score, Perplexity, Change and Success Rate
    & Pattern search by human experts\\

    \textbf{Knowledge Illusion} \cite{DBLP:journals/corr/abs-2104-07143} & 
    %Local (Set of Individual Neurons) &
    %Lexical properties & 
    Lexical, Geometric Properties (Local Semantic Coherence) & 
    Next Sequence Prediction &
    Projection Score, Activation Quantile, Word Frequency Correlation
    & Human annotations for patterns using visualization\\
       
    \bottomrule
     \end{tabular}
      
\end{table*}

\section{Related Surveys}

As the interest in the Explainable Artificial intelligence (XAI) field grows, various survey articles were published, trying to consolidate and categorize the approaches. We segregate the reviews into two categories: Surveys that give a general overview of existing explainability methods \cite{Adadi2018PeekingIT, 10.1145/3236009, DBLP:journals/corr/abs-1909-03012, e23010018, 9803681} and surveys that focus on explainability methods in the NLP domain. We narrow our surveys to the NLP domain as this is the core focus of this review paper. 

A survey that acts as a prior to ours is from \citet{10.1162/tacl_a_00254}, where the authors review the various analysis methods used to conduct novel and fine-grained neural network interpretation and evaluation. The primary question that has been relevant while formulating these interpretation methods is: What linguistic information is captured in neural networks? The authors emphasize three aspects of the language-specific analysis, namely, methods used for conducting the analysis, linguistic information sought, and neural network parts investigated. They also identify several gaps and limitations in the surveys. 

\citet{danilevsky-etal-2020-survey} presents a broader overview of the state of XAI over a span of 7 years (until 2020), with a focus on the NLP domain. This work focuses on outcome explanation problems which help end users understand the model's operation and thereby build trust in these NLP-based AI systems. Along with the high-level classification of explanations, the work introduces two additional aspects: techniques that derive the explanation and techniques to present to the end user. The explainability techniques are categorized into feature importance, surrogate models, example-driven, provenance-based and declarative induction. A set of operations such as first-derivative salience, layer-wise relevance propagation, input perturbations, attention mechanism, and Long-Short-Term-Memory (LSTM) gating signal and explainability-aware architectures enable explainability. An interesting observation is the consideration of adding attention layers to neural network architectures as a strategy to enable explanations.

The closest survey related to our work is from \citet{DBLP:journals/corr/abs-2108-13138}, where the survey is on fine-grained neuron analysis. While there have been two previous surveys that cover Concept Analysis \cite{DBLP:journals/corr/abs-1911-00317} and Attribution Analysis \cite{danilevsky-etal-2020-survey}, their focus is on analyzing individual neurons to better understand the inner workings of neural networks. They refer to this as Neuron Analysis and categorized these reviewed methods into visualization, corpus-based, neuron probing, and unsupervised methods. The work further discusses findings and applications of neuron interpretation and summarizes open issues. 

We observe that, from the various existing surveys, there are different dimensions to be considered. We narrow down our survey into the following dimensions: 
\begin{itemize}
    \item \textit{Analysis methods} that focus on the internal interpretation of the activation space.
    \item \textit{Linguistic Information} such as parts-of-speech, syntactic, semantic and \textit{Non-linguistics Information} such as sentence length, factual knowledge, geometric properties.
    \item \textit{Neural network object} neurons and its activations as the Activation Space in the transformer language model.
\end{itemize}

We believe that interpretability alone is not sufficient in understanding the inner workings of the transformers, we also need explainability to summarize the reason for the model's behaviour in a human-comprehensible manner. One has to keep in mind that, explainability and interpretability have distinguishable meanings \cite{DBLP:journals/corr/abs-1806-00069} and our review focuses only on interpretability methods because the research works reviewed focus on the same.

\section{Survey Methodology}

Our survey aim to cover the advances in NLP XAI research focusing on neuron interpretation. As defined earlier, we define this latent dimension as \textit{Activation Space} and consider the reviewed techniques as \textit{Activation Space Analysis} methods. We filtered to those methods that work at the feed-forward neuron-level, individual vs global, within the transformer model. We identified relevant papers published in NLP and AI conferences (AAAI, ACL, IJCNLP, EMNLP) between 2018 and 2022. With the limited scope of neuron-level analysis, we arrived at seven contemporary papers. With a limited number of work in this direction, we decided to take a deeper look into each of these methods, analyze its benefits, limitations, and gaps and present this study as our review paper. We are aware that this is an ongoing and relatively new research field and our focus is extremely limited; we acknowledge that we might have omitted certain papers. We also assume that if the authors have focused on explainability, they are more likely to cover the relevant related taxonomies, categories, and methods. Another common observation is that explanations are generated in an NLP task-oriented setting and remain relevant to the task context. Even though we summarize the tasks on which these researches are based, the task definitions are not relevant in our review process of understanding the activation space. 

\section{Taxonomies and Categorization}

There still exists a reasonably vague understanding and lack of concrete mathematical definition between the two commonly used terms: explainability and interpretability. Interpretability has been defined as "the degree to which a human can understand the cause of a decision" \cite{DBLP:journals/corr/Miller17a} or the degree to which a human can consistently predict the model’s result \cite{NIPS2016_5680522b}. A broader definition exists for the term \textit{interpretable machine learning} as the extraction of relevant knowledge from a machine-learning model concerning relationships either contained in the data or learned by the model. This definition rather focuses on understanding what the model learns either from an input-output mapping perspective or what the model itself learns. 
%The contrast between the above two definitions is that the former focuses on understanding the decision-making process of the model whereas the latter is concerned with extracting the relevant knowledge.  
On the other hand, explainability directs the focus back to human understanding by examining the relationship between input features and model predictions in a human-understandable format \cite{e23010018}.

After reviewing numerous relevant existing literature, we observed that explainability techniques broadly fall into three major classes. The first differentiates between understanding a model's  individual prediction process versus prediction process as a whole \cite{danilevsky-etal-2020-survey}. A second differentiation is made in self-explaining or post-hoc methods, where the former generates explanations along with the model’s prediction process whereas the latter requires post-processing of elements extracted during the model prediction process. The third major distinction corresponds to methods that are model specific or agnostic in nature. We also observed the existence of various other categorizations like outcome-based explanations, visual explanation methods, operations, and conceptual vs attribution. Visualization methods play a salient role in further understanding any interpretation method \cite{pezeshkpour-etal-2019-investigating,mullenbach-etal-2018-explainable, croce-etal-2019-auditing, https://doi.org/10.48550/arxiv.1409.0473}. These methods are inherent to interpretability and is been widely reviewed, we leave this to the reader to explore the relevant literature. 

\section{Activation Space Analysis Methods}
\label{sec:aaa}
There are two types of interpretability analysis that are carried out in the related research work: 1) Analyze individual neurons and 2) Analyze the entire set of neurons of the feed-forward layer. We look into both approaches from four perspectives: categorization, linguistic knowledge sought for, methodology, and evaluations, and conduct a comparative analysis of these methods.

%\subsection{Linguistic Phenomena}

\textbf{Linguistic Phenomena:} Investigating the linguistic phenomena that occurs within the activations of pre-trained models, when trained for a specific task set, using various interpretability analysis methods, is a common way to interpret the features learned by these models. The linguistic phenomenon refers to the presence of various linguistic features such as word morphology, lexical semantics, syntax or linguistic knowledge such as parts-of-speech, grammar, coreference, lemmas. Linguistic Correlation Analysis (LCA) is one such method that focuses on understanding what the model learned about linguistic features and determining those neurons that explicitly focus on such phenomena.  A toolkit with three major methods, Individual Model Analysis, Cross-model Analysis and LCA, to identify salient neurons within the model or related to a task under consideration, is presented by \citet{dalvi2019neurox}. 
%The three major methods in here are: 1) Individual Model Analysis that enables visualization of individual neurons and improvement in architectural design choices thereby comprehending model quality, 2) Cross-model Analysis that determines the correlation-based neuron ranking across several models, allowing to understand the properties that the models learn specific to the task, and 3) Linguistic Correlation Analysis (LCA)

Probing using diagnostic classifiers to understand the knowledge captured in neural representations is another common method for associating model components with linguistic properties \cite{https://doi.org/10.48550/arxiv.1711.10203, DBLP:journals/corr/abs-1805-01070, DBLP:journals/corr/abs-1812-08951}. This involves extracting feature representations from the network and training an auxiliary classifier to predict the linguistic property. Layer-wise and neuron-level diagnostic classifiers that probe representation from individual layers w.r.t linguistic properties and find neurons that capture salient features, respectively, are used to conduct analysis on pre-trained models BERT, RoBERTa and XLNet \cite{durrani-etal-2021-transfer}. %Even though this research is a detour from our core focus of this review work, we still decided to include it for the reason that the study focuses on understanding linguistic features at the neuron level for individual transformers and linguistic information distribution across the models.
The task of predicting a certain linguistic property is defined. A diagnostic classifier (logistic regression) is trained on generated activations, for both layer-wise and neuron-wise probes, to predict the existence of this linguistic property. % The classifier used is logistic regression for both layer-wise and neuron-wise probes and minimizes a loss function that depends on the word probability in the input dataset that gets assigned to the linguistic property under consideration. 
An LCA is conducted to generate neuron ranking based on weight distribution. Additionally, an elastic-net regularization is fine-tuned using grid-search to balance between focused and distributed neurons. The top N salient neurons extracted from this ranked list are used to retrain the classifier until an Oracle accuracy is achieved.

\citet{DBLP:journals/corr/abs-2010-02695} and \citet{alammar-2021-ecco} conducts similar experiments, where the entire neuron activations from the feed-forward layers are used to train an external classifier. \citet{DBLP:journals/corr/abs-2010-02695} uses a probing classifier (logistic regression) with the additional elastic-net regularization to conduct a fine-grained neuron level analysis on pre-trained models ELMo, T-ELMo, BERT, and XLNET. This variance of models, in this study, covers different modeling choices of the blocks, optimization objectives, and model architectures. The case study conducted by \citet{alammar-2021-ecco} uses probing the feed-forward neuron activations for Parts-of-Speech (POS) Information. A control task is created where each token is assigned to a random POS tag and a separate probe is trained on this control set. This allows us to measure the difference in prediction accuracy between the actual and control dataset, selectivity score, thereby concluding if the probe really extracts the POS information. %This difference in accuracy is referred to as the Selectivity score.
The author collects existing methods that examines input saliency, hidden state evolution, neuron activations, and non-negative matrix factorization of neuron activations, along with dimensionality reduction methods to extract patterns into an open-source library known as Ecco \cite{alammar-2021-ecco}. These methods can be directly employed on pre-trained models such as GPT2, BERT, RoBERTa.

%The author collects existing methods used to examine neuron activations in the NLP and computer vision domain, along with additional dimensionality reduction methods to extract patterns into an open-source library known as Ecco \cite{alammar-2021-ecco}. It provides a broad set of tools that examines input saliency, hidden state evolution, neuron activations, and non-negative matrix factorization of neuron activations and can be directly employed on pre-trained models such as GPT2, BERT, RoBERTa.

%\subsection{Neural Memory Cells}
\textbf{Neural Memory Cells:} In the context of a neural network with a recurrent attention model, \citet{https://doi.org/10.48550/arxiv.1907.01470} introduced input and output memory representations. A recent work extends this neural memory concept and shows that the feed-forward layers in the transformer models operate as key-value memories, where keys correlate to specific human-interpretable input pattern sets and simultaneously, values induce a distribution over the output vocabulary \cite{DBLP:journals/corr/abs-2012-14913}. The work analyzes these memories present in the feed-forward layers and further explores the function of these layers in transformer-based language models. 

A neural memory is defined as a key-value pair, where each key value is a \textit{d}-dimensional vector. %The mathematical definitions of the feed-forward layer and key-value neural memories are shown to be identical with the only difference being that neural memory uses softmax as the non-linearity whereas the feed-forward layer in the transformer does not have such a normalizing function. 
The emulation, mathematical similarity between feed-forward and key-value neural memories, allows the hidden dimension to be considered as number of memories in each layer and the activations as vectors containing un-normalized non-negative memory coefficients. Using this similarity, the study posits that the key vectors act as pattern detectors. This hypothesis is tested by looking for the highest memory coefficient that is associated with the input text, retrieving those input examples, and conducting human evaluations to identify patterns. The study further explores intra-layer memory composition and inter-layer prediction refinement.

The concept of knowledge neurons, neurons that express a fact, is introduced by \citet{DBLP:journals/corr/abs-2104-08696}. The authors propose a method to find the neurons that express facts and how their activations correlate in expressing these facts. The evaluations on pre-trained models for fill-in-the-blank cloze tasks show that these models have the ability to recall factual knowledge even without fine-tuning. The work considers feed-forward layers as key-value memories, hypothesize that these key-value memories store factual knowledge and proposes a knowledge attribution method. The knowledge attribution method, based on integrated gradients, evaluates the contribution of each neuron, in BERT-base-cased transformer, to knowledge predictions by assigning them an attribution score. Those neurons with a higher gradient i.e attribution score are identified as those contributing to factual expressions. Further refinement of these neurons is done under the hypothesis that there are chances that the same fact can share the same set of true positive knowledge neurons. This refinement allows in retaining only those knowledge neurons that are shared by a certain percentage of input prompts.

%They hypothesize that these key-value memories store factual knowledge. The work experiments on BERT-base cased transformer and further studies the effect of manipulating these extracted knowledge neurons in prediction accuracy.

%A fact is defined as a relational triplet, consisting of head and tail entities and the relation between them. The cloze query requires the pre-trained model to answer the query but leaves the tail entity blank. Such queries are referred to as knowledge-expressing prompts and these queries are used to find the specific knowledge neurons. The Knowledge attribution method, based on integrated gradients, evaluates the contribution of each neuron to knowledge predictions by assigning them an attribution score. Those neurons with a higher gradient i.e attribution score are identified as those contributing to factual expressions. Further refinement of these neurons is done under the hypothesis that there are chances that the same fact can share the same set of true positive knowledge neurons. This refinement allows us to retain only those knowledge neurons that are shared by a certain percentage of prompts.

%\subsection{Knowledge Illusion}
\textbf{Knowledge Illusion:} Based on the generalization of the hypothesis that concepts are encoded in the linear combinations of neural activations, \citet{DBLP:journals/corr/abs-2104-07143} describe a surprising phenomenon ``interpretability illusion''. Probing experiments conducted on BERT-base-uncased model determines if individual neurons contained human-interpretable meaning. The final layer creates embeddings for four datasets (QQP, QNLI, Wiki, and Books) and 
%the top activating sentences by sorting sentence embeddings according to their activation level are determined. 
top 10 activating sentences for a neuron are annotated to determine a pattern. Here a pattern is defined as a single property such as sentence length or lexical similarity shared by a set of sentences. By proposing three sources: dataset idiosyncrasy, local semantic coherence in BERT’s embedding space, and annotator error, the authors explain this illusion. The same experiment is repeated, by keeping a set of target neurons constant, on various datasets to reveal the illusion as described by the authors. The work further explores the causes of this illusion by investigating local, global and dataset-level concepts. 

%As mentioned earlier there is limited research done in analyzing neuron activation. %LCA and Ecco are tool kits that facilitate further analysis and in our view can be a good starting point to conduct further activation space analysis research.  In the next section, we present the various evaluation techniques employed in the above-presented methods, identify the gaps present and indicate our observations. 
 %Note that we have chosen an indicative name for each method by considering the most common representative keyword from the respective papers.

\section{Evaluations}

\textbf{Linguistic Phenomena:} A layer-wise probing is conducted to understand the redistribution of linguistic knowledge (syntactic chunking, POS, and semantic tagging) when fine-tuned for downstream tasks \cite{durrani-etal-2021-transfer}. Using this probing across three fine-tuned models BERT, RoBERTa, and XLnet, on GLUE tasks and architectures reveal the following observations: The morpho-syntactic linguistic phenomenon that is preserved, post fine-tuning, in the higher layers is dependent on the task; Different architectures preserve linguistic information differently post fine-tuning. The neuron-wise probing further refines to the fine-grained neuron level, where the most salient neurons are extracted and their distribution across architecture and variations in downstream tasks are studied. An alignment of findings is found with \citet{merchant-etal-2020-happens}, where the fine-tuning affects only the top layer. In comparison with \citet{mosbach-etal-2020-interplay-fine}, which is focused on sentence level probing, \citet{durrani-etal-2021-transfer} studies core-linguistic phenomena. Additionally, their findings from fine-grained neuron analysis extend the core-linguistic task layer-wise analysis, along with fine-tuning effects on these neurons. Another interesting observation made is the different patterns that are entailed when these networks are pruned from top or bottom. 
%A deeper study with wider architecture considerations on four pre-trained models ELMo, Transformer-ELMo, BERT, and XLNet on four NLP tasks: syntactic chunking, POS, semantic and syntax tagging is conducted by \citet{DBLP:journals/corr/abs-2010-02695}. 

An ablation study conducted by \citet{DBLP:journals/corr/abs-2010-02695} on the top salient neurons, from four pre-trained models ELMo, T-ELMo, BERT, and XLNet, indicates higher distribution of linguistic information across the network when the underlying task is more complex (CCG supertagging), revealing information redundancy. Further refined study, considering only a minimal set of neurons, to identify the network parts that predominantly capture the linguistic information and understand the localization or distribution of this information, indicate that the number of neurons required to achieve the Oracle accuracy varies and is dependent on the complexity of the task. By employing a selectivity score next to the prediction accuracy score, and training separate POS probes for the actual dataset and a control task, \citet{alammar-2021-ecco} observes that the activation space encodes POS information at levels comparable to BERT's hidden states. The non-negative matrix factorization method helps in identifying those patterns in neuron activations that correspond to syntactic and semantic properties of the input text. The NeuroX toolkit is compared with the What-if tool from Google, that inspects trained models based on prediction and Seq2Seq-Vis \cite{DBLP:journals/corr/abs-1804-09299}, that can trace back prediction decisions in Neural Machine Translation input models \cite{dalvi2019neurox}. 

\textbf{Neural Memory Cells:} Relating the patterns identified by human experts (NLP graduate students) to human understanding, the patterns are classified as shallow or semantic and are associated with lower layers and upper layers of a 16-layer transformer model, respectively \cite{DBLP:journals/corr/abs-2104-08696}. Further analysis of the corresponding values from the key-value memories complements the patterns observed in the respective keys. The agreement rate, the fraction of memory cells that match the corresponding keys and values, is seen to increase in higher layers. The authors suggest that the memory cells in the higher layers contribute to the output whereas the lower layers do not show such a clear key-value correlation to contribute toward the output distribution of the next word. A qualitative analysis, by manually analyzing a few random cases, is conducted on the layer-wise distribution of memory cells and how the model refines its prediction from layer to layer using residual connections. The work is an extension of \citet{https://doi.org/10.48550/arxiv.1907.01470}, which suggests a theoretical similarity between feed-forward layers and key-value memories. Additionally their observations, of shallow feature encoding, confirms with recent findings from \citet{peters-etal-2018-deep, jawahar-etal-2019-bert, liu-etal-2019-linguistic}.

The BERT-base-cased model is experimented with the knowledge attribution, where activation value is considered as the attribution score for a neuron, to measure neuron sensitivity towards input. Similar observations to \citet{DBLP:journals/corr/abs-2012-14913} and \citet{tenney-etal-2019-bert} are identified: fact-related neurons are distributed in the higher layers of the transformer. Further, the authors investigate how these neurons contribute to expressing the knowledge either by suppressing or amplifying their activations. Two additional use cases, updating facts and erasing relations, are presented, where the authors demonstrate the potential application of these identified knowledge neurons. Two evaluation metrics are used: change and success rate for measuring fact updating and inter/intra-relation perplexity for measuring the influence on other knowledge. These evaluations indicate that changes in very few neurons in the transformers can affect certain facts. Erasing of facts is also measured using perplexity and is observed that post fact erasing operation, i.e. setting knowledge neuron to zero vectors, the perplexity of the moved knowledge increased. The knowledge attribution method, built on integrated gradients, is inspired by \citet{https://doi.org/10.48550/arxiv.2004.11207} and \citet{pmlr-v70-sundararajan17a}.

\textbf{Knowledge Illusion:} A qualitative evaluation is conducted by annotating three sets of sentences for a neuron in consideration: 1) top ten activating sentences for the neuron, 2) top ten activating sentences in random direction and 3) ten random sentences \cite{DBLP:journals/corr/abs-2104-07143}. The objective of this annotation is to find patterns, where a pattern is defined as a property shared by a set of sentences. A pattern is considered as a proxy for a learned concept by the model. For each neuron under consideration, an average of 2.5 distinct patterns across four datasets are observed. This illusion is further explored by studying the regions of activation space the input data occupies, the influence of top activating sentences on patterns from both local semantic coherence and global directions, and annotation error. Qualitative analysis is conducted through (UMAP dimensionality reduction) visualization and it is observed that sentences cluster in accordance with datasets. Additionally, the high accuracy of a Support Vector Machine classifier distinguishes between these datasets and provides quantitative evidence for this observation. This indicates the dependence of information encoded within neurons on the idiosyncrasies of the natural language datasets, even though they have similar activation values. The analysis of global directions in BERT’s activation space using activation quantiles helps in understanding the correlation between word frequency change and its monotonicity in each combination of datasets. This correlation indicated that despite BERT's illusionary effect, there still exists meaningful global direction in its activation space.  While comparing the observed illusions with previous works, it is in alignment with \citet{aharoni-goldberg-2020-unsupervised}, where they demonstrate the usage of BERT representations to disambiguate datasets. This explains the existence of patterns in datasets, further experiments are conducted to understand the cause of such pattern existence. 

We observe that all the methods that we reviewed so far fall under the local interpretability methods and limit themselves to the top N salient neurons (see Table \ref{tab:commands}). From reviewing these studies, we observe dimensionality reduction is required to understand the properties under consideration. Dimensionality reduction is associated with information loss and this loss is not accounted for in these studies. Another observation is that the focus of these studies alternates between identifying the neurons that capture the relevant linguistic information and those subsets of these neurons that affect the prediction accuracy. Moreover, some interpretability methods are evaluated through user studies (where users subjectively evaluate the explanations), whereas others are evaluated in terms of how they satisfy some properties, either quantitatively or qualitatively, without real users’ evaluations. In the next section, we further discuss our observations and present our insights and future detections.

\section{Insights and Future Directions}
\label{sec:ins}
A common observation that we see in the contemporary general surveys and from our focused reviews is the lack of both theoretical foundations and empirical considerations in evaluations \cite{DBLP:journals/corr/abs-2108-13138, 10.1162/tacl_a_00254, danilevsky-etal-2020-survey}. Even though each method has quantitative measures for evaluation, there is no standard set of metrics for comparing various observations, hence, confining the scope of respective interpretability technique results to specific model architectures or task-related domains. Studies have proposed various desiderata for interpretable concepts such as Fidelity, Diversity and Grounding \cite{NEURIPS2018_3e9f0fc9} for qualitative consistency %but the studies do not conform to these properties in their qualitative evaluation. 
Additionally, a few studies employ human experts for qualitative analysis such as pattern annotation and identifications, but again lack a standard framework for a comparative study and consistent explanations. Moreover, the subjective nature of interpretability and the lack of existence of ground truth in qualitative analysis makes it even more challenging to evaluate these methods. 

By reviewing the above works, that focus on activation space, we observe the following from the model perspective: For a fixed model architecture and when a fixed set of neurons are examined, each set of neurons encode different information, dependent on the input dataset; On the contrary, when a wider set of model architectures are considered, the same set of neurons encode similar information at lower and higher layers across these architectures but the information encoded is dependent on the underlying task. These observations emphasize the dependency on the input data and the underlying task of interpreting the linguistic information encoded in the activation space. 

Experiments conducted align with the definition of interpretability and explainability in understanding the rationale behind the model's decision but lack human understandable explanations. In the context of explainability, we observe that there is a gap in human-understandable linguistic concepts and linguistic features captured in the network. %The existing methods investigate the linguistic features captured both locally and globally in the model.
We make a clear distinction between linguistic features and concepts: features consist of linguistic properties such as parts-of-speech, syntactic and semantic properties, and word morphology whereas the linguistic concepts, from a human understandable perspective, encode general human knowledge and how it is expressed in natural language. 
%We believe the knowledge representation available in ConceptNet \cite{speer2017conceptnet,speer-havasi-2012-representing}, which expresses concepts such as color, objects, as words and phrases from natural language is a useful conceptual representation technique. This technique that can be used as a baseline while seeking conceptual information captured in the activation space. In this manner, the gap between human understandable linguistic concepts and linguistic features captured in the models can be diminished.  
Various contemporary methods such as Concept Relevant Propagation \cite{https://doi.org/10.48550/arxiv.2206.03208}, Testing Concept Activation Vector \cite{https://doi.org/10.48550/arxiv.1711.11279}, Integrated Conceptual Sensitivity \cite{DBLP:journals/corr/abs-2106-08641} that are based on human understandable local and global concept-based explanations exist. These methods are applied and evaluated in the image processing domain and are yet to be explored in understanding linguistic concepts. It is evident that exploring activation space is a promising research direction and we propose a potential future direction: extend the interpretability techniques from image processing to the natural language processing domain through transfer learning.

\begin{acknowledgments}
  The authors would like to thank the anonymous reviewers for their helpful feedback. The work was partially funded by the German Federal Ministry of Education and Research (BMBF) through the project XAINES (01IW20005). 
\end{acknowledgments}

%%
%% Define the bibliography file to be used
\bibliography{aimlaiws}

%%
%% If your work has an appendix, this is the place to put it.
\appendix
\section{Evaluation Metrics Definitions} 
\begin{itemize}
    \item \textit{Selectivity:} The difference between linguistic task accuracy and control task accuracy
    \item \textit{Prediction Accuracy:} Performance measure of the model on a given task
    \item \textit{Agreement Rate:} The fraction of memory cells (dimensions) where the value’s top prediction matches the key’s top trigger example
    \item \textit{Value Probability:} Probability of the values' top prediction
    \item \textit{Projection Score:} The dot product between a sentence embedding and a direction
    \item \textit{Activation Quantile:} Equally sized smaller subsection of the activation space
    \item \textit{Word Frequency Correlation:} The correlation between directions and words in the embedding space
    \item \textit{Attribution Score:} Measures the contribution of the neuron to the factual expressions 
    \item \textit{Perplexity:} Measurement of how well a probability model predicts a sample, degree of ‘uncertainty’ a model has in predicting
    \item \textit{Change Rate:} The ratio that the original prediction is modified to another
    \item \textit{Success Rate: } The ratio that becomes learned prediction the top predictions
\end{itemize}

%\section{Online Resources}

%The sources for the ceur-art style are available via
%\begin{itemize}
%\item \href{https://github.com/yamadharma/ceurart}{GitHub},
% \item \href{https://www.overleaf.com/project/5e76702c4acae70001d3bc87}{Overleaf},
%\item
%  \href{https://www.overleaf.com/latex/templates/template-for-submissions-to-ceur-workshop-proceedings-ceur-ws-dot-org/pkfscdkgkhcq}{Overleaf
%    template}.
%\end{itemize}

\end{document}